\documentclass[letterpaper, 10 pt, conference]{ieeeconf}  

\IEEEoverridecommandlockouts                              

\usepackage{graphics} 
\usepackage{epsfig} 
\usepackage{mathptmx} 
\usepackage{times} 
\usepackage{amsmath} 
\usepackage{amssymb}  
\usepackage{multirow, bigdelim}
\usepackage{amsfonts}
\usepackage{booktabs}
\usepackage{threeparttable}
\usepackage[table,xcdraw]{xcolor}
\usepackage{cite}
\usepackage{}
\usepackage{url}
\usepackage{algorithmic}
\usepackage[ruled,vlined,linesnumbered]{algorithm2e}
\usepackage[pagebackref=true,breaklinks=true, colorlinks, bookmarks=false]{hyperref}

\title{\LARGE \bf
 PANet: LiDAR Panoptic Segmentation with Sparse Instance \\ Proposal and Aggregation
}

\author{
Jianbiao Mei$^{1,*}$, Yu Yang$^{1,*}$, Mengmeng Wang$^{1}$, Xiaojun Hou$^{1}$, Laijian Li$^{1}$ and Yong Liu$^{1,\dag}$
\thanks{
Jianbiao Mei* and Yu Yang* contribute equally.
$^{1}$The authors are with the Institute of Cyber-Systems and Control, Zhejiang University, Hangzhou, 310027, China. (Yong Liu$\dag$ is the corresponding author, email: yongliu@iipc.zju.edu.cn)
}}%

\begin{document}

\maketitle
\thispagestyle{empty}
\pagestyle{empty}

\begin{abstract}
Reliable LiDAR panoptic segmentation (LPS), including both semantic and instance segmentation, is vital for many robotic applications, such as autonomous driving. This work proposes a new LPS framework named PANet to eliminate the dependency on the offset branch and improve the performance on large objects, which are always over-segmented by clustering algorithms. Firstly, we propose a non-learning Sparse Instance Proposal (SIP) module with the ``sampling-shifting-grouping" scheme to directly group thing points into instances from the raw point cloud efficiently. More specifically, balanced point sampling is introduced to generate sparse seed points with more uniform point distribution over the distance range. And a shift module, termed bubble shifting, is proposed to shrink the seed points to the clustered centers. Then we utilize the connected component label algorithm to generate instance proposals. 
Furthermore, an instance aggregation module is devised to integrate potentially fragmented instances, improving the performance of the SIP module on large objects. Extensive experiments show that PANet achieves state-of-the-art performance among published works on the SemanticKITII validation and nuScenes validation for the panoptic segmentation task. Code is available at \url{https://github.com/Jieqianyu/PANet.git}.
\end{abstract}

\section{Introduction} \label{Sec.intro}
LiDAR Panoptic Segmentation (LPS) plays an important role in 3D scene understanding using point clouds, which has been an essential task for many robotic applications such as autonomous driving. LPS combines semantic and instance segmentation in a single framework, providing both semantic labels for points in the scenes and instance IDs for points that belong to instances (things). With the emergence of large-scale point cloud benchmarks, e.g., SemanticKITTI \cite{behley2019semantickitti} and nuScenes \cite{fong2022panoptic}, LPS has achieved rapid progress. However, performing reliable panoptic segmentation is still highly challenging due to the sparse, unordered, and non-uniform sampled natures of point clouds.

Existing methods for LPS can be mainly divided into 
detection-based \cite{hurtado2020mopt, sirohi2021efficientlps, xu2023aop, ye2022lidarmultinet} and clustering-based \cite{milioto2020lidar, li2022smac, hong2021lidar, zhou2021panoptic, gasperini2021panoster, liu2022prototype, duerr2022rangebird, li2022PHNet} approaches.
The former applied the 3D object detection network to discover instances, which are usually limited by detection accuracy. 
The latter achieved instance segmentation through the center regression and clustering algorithms. For clustering-based methods, the clustering performance is easily affected by the point distribution of the regressed centers. And we found that most existing methods heavily depend on the learnable offset branch to provide geometric shifts for center regression. However, it is hard to predict ideal geometric shifts due to the sparsity, non-uniform density of LiDAR point cloud, and various shapes/sizes of instances. Recently, DSNet \cite{hong2021lidar} designed a learnable dynamic shifting (DS) module to further shift the regressed centers to the clustering centers iteratively, but the shifted centers may not match the ground-truth instance centers. The possible inconsistency degrades the learning of the DS module.

To address the above problems, we propose a non-learning Sparse Instance Proposal (SIP) module to directly group instances from the raw points of things efficiently. We adopt the ``sampling-shifting-grouping" scheme to design our SIP module. Specifically, to avoid the significant computational burden and memory overhead caused by shifting and grouping all the points of things, we introduce a balanced point-sampling (BPS) strategy. The proposed BPS generates sparse seed points with a more uniform distribution over the distance range and implements the point sampling and assignment simultaneously, improving the efficiency. Furthermore, a simple but effective shift module, termed Bubble Shrinking (BS), is devised to efficiently and precisely shift the seed points to the clustered centers iteratively. Finally, we group the shifted points into instances by the Connected Component Labeling (CCL) algorithm, which can be implemented by the efficient depth-first search.
Due to the cascade design of BPS, BS, and CCL-based grouping, our non-learning SIP module is effective and efficient and can be easily extended to other backbones and datasets in a plug-and-play manner.

Nevertheless, due to the sparsity of LiDAR point clouds, there are cases where fragmented/over-segment instances may be generated by clustering algorithms, including our SIP module, especially when grouping large objects such as trucks and buses. 
To improve the completeness of instance segmentation, we propose an Instance Aggregation (IA) module to further integrate the potentially fragmented instances. More specifically, we apply the KNN-Transformer to enhance the interactions among the instance proposals and merge the instance proposals that belong to the same instance ID by the instance affinities. Our IA module can complement the SIP module, further improving the segmentation performance of large objects. 

Extensive experiments on two large-scale datasets SemanticKITTI \cite{behley2019semantickitti} and nuScenes \cite{fong2022panoptic} demonstrate the effectiveness of our method PANet. Our contributions are summarized as follows:

$\bullet$ We develop a Sparse Instance Proposal (SIP) module without extra learning tasks to directly group instances from the raw points of things, which can be easily extended to other backbones and datasets in a plug-and-play manner. 

$\bullet$ SIP eliminates the dependency on the offset branch and accelerates the clustering process due to the cascade design of BPS, BS, and CCL-based grouping.

$\bullet$ We propose an instance aggregation module to integrate the possible fragmented instances and complement the SIP module to improve large objects' segmentation performance.

$\bullet$ Extensive experiments on both SemanticKITTI \cite{behley2019semantickitti} and nuScenes \cite{fong2022panoptic} datasets show that our model achieves state-of-the-art performance.

\section{Related Work}
\subsection{LiDAR Semantic Segmentation.}
According to the data representations, most semantic segmentation methods on point clouds can be categorized into projection-based, point-based, voxel-based, and multi-view methods. Projection-based works \cite{cortinhal2020salsanext, milioto2019rangenet++, zhang2020polarnet, xu2020squeezesegv3, gu2022maskrange} project the raw point clouds into a certain plane such as range-view and BEV, which benefit from some efficient 2D CNN architectures but do not always reflect 3D relationships. 
Following PointNet/PointNet++ \cite{qi2017pointnet, qi2017pointnet++}, point-based methods \cite{thomas2019kpconv, hu2020randla} directly process the raw point clouds, which, however, take a time-consuming local neighborhood search.
Voxel-based methods \cite{zhu2021cylindrical, ye2021drinet, ye2022efficient} voxelized the raw point clouds and usually apply 3D sparse convolutions \cite{graham20183d} to extract voxel-wise features, reducing the computational burden and memory overhead.
Multi-view methods \cite{duerr2022rangebird, xu2021rpvnet, gerdzhev2021tornado} fuse different representations of point clouds to exploit their individual properties.
Similar to \cite{li2022PHNet}, we combine a sparse 3D CNN and a tiny 2D U-Net to aggregate multi-scale 3D features and 2D features to improve semantic segmentation, a vital part of panoptic segmentation.
\subsection{LiDAR Panoptic Segmentation}
Most LiDAR panoptic segmentation (LPS) methods usually consist of semantic and instance branches. Regarding the implementation of instance segmentation, these approaches can be classified into detection-based and clustering-based methods. 

\textbf{Detection-based methods.} These methods \cite{hurtado2020mopt, sirohi2021efficientlps, xu2023aop, ye2022lidarmultinet} integrate 3D object detection \cite{yin2021center, zhou2022centerformer, fan2022embracing, lang2019pointpillars} to discover instances. 
They adopt the detector to explicitly provide instances' location and size information for further segmentation. 
PanopticTrackNet \cite{hurtado2020mopt} utilizes Mask R-CNN \cite{he2017mask} for instance segmentation. 
EfficientLPS \cite{sirohi2021efficientlps} fuses the semantic logits, bounding boxes, and mask logits and generates the panoptic segmentation results in the range view, which are further back-projected to obtain final predictions. AOP-Net \cite{xu2023aop}, and LidarMultiNet \cite{ye2022lidarmultinet} designs a multi-task pipeline to combine 3D object detection and panoptic segmentation.
These methods benefit from the prior information provided by the detector, but the segmentation performance largely depends on the detection results.

\begin{figure*}[t]
\centering
	\includegraphics[width=0.9\textwidth]{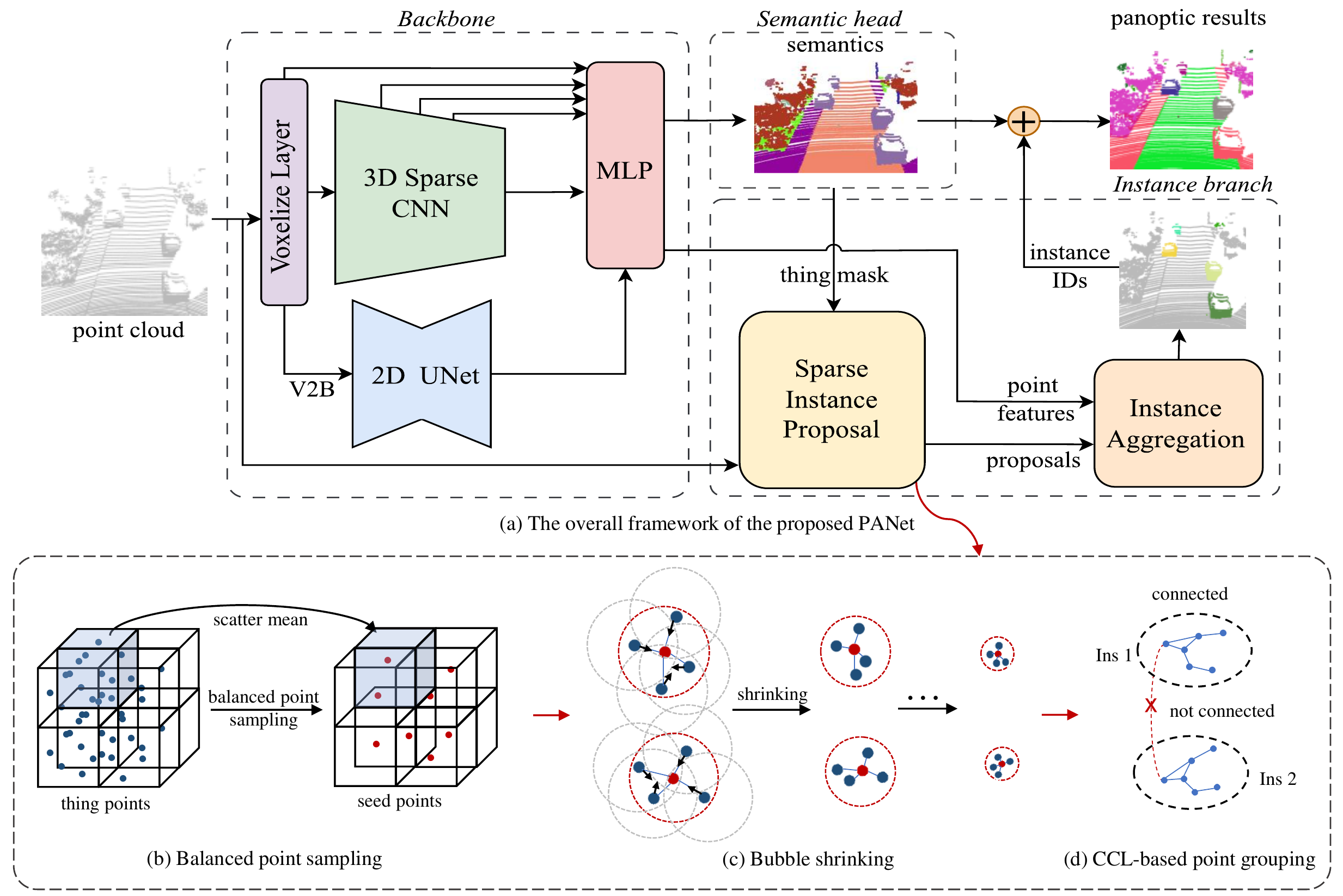}
	\caption{The overview of the proposed PANet. The backbone consists of a multi-scale sparse 3D CNN and a tiny 2D U-Net to aggregate multi-level 3D and 2D features. The extracted features are fed into the semantic head for semantic predictions. In the instance branch, the Sparse Instance Proposal (SIP) module shown in the bottom half is proposed to efficiently group the raw points of things into instances. Moreover, the Instance Aggregation (IA) module takes point-wise features and integrates the potentially fragmented instances generated by the SIP module. Finally, the semantic predictions and merged instances are combined to obtain the final panoptic segmentation results. ``V2B" denotes the projection of voxel features to BEV features.}
	\label{fig:pipeline}
\end{figure*}

\textbf{Clustering-based methods.} These methods \cite{milioto2020lidar, li2022smac, hong2021lidar, zhou2021panoptic, gasperini2021panoster, liu2022prototype, duerr2022rangebird, li2022PHNet, razani2021gp} perform clustering algorithms to group thing points into instances. Panoptic-PolarNet \cite{zhou2021panoptic} predicts the 2D center heatmap and points shifts for clustering. DS-Net \cite{hong2021lidar} designed a learnable dynamic shift module to further shift the regressed centers to the clustering centers. Panoptic-PHNet \cite{li2022PHNet} introduced a pseudo heatmap generated from the shifted thing points and a center grouping module to yield instance centers for efficient clustering. There are also studies \cite{razani2021gp} over-segmenting the instances and utilizing graph networks to aggregate the fragments. We also devise a clustering scheme. Unlike these methods, the proposed sparse instance proposal module without extra learning tasks eliminates the dependency on the offset head. It allows us to extend it to other backbones and datasets in a plug-and-play manner. Moreover, different from \cite{razani2021gp}, which merges a large amount of artificially over-segmented proposals with GNN, our instance aggregation module focuses on improving the SIP module's performance on large objects and utilizes the KNN-Trasformer to model the relationships among sparse potentially fragmented instances, which is more effective and efficient.

\section{Method}
\subsection{Backbone Design} \label{Sec.BD}
Similar to Panoptic-PHNet \cite{li2022PHNet}, we combine a multi-scale sparse 3D CNN and a tiny 2D U-Net to aggregate multi-scale 3D features and 2D features. As Fig.\ref{fig:pipeline}(a) shows, the input LiDAR point cloud $P \in \mathbb{R}^{N\times4}$ (coordinates and intensity) is first fed into a voxelization layer (similar to the voxelization in DRINet \cite{ye2021drinet}) to obtain the voxel-wise features $F^0_v \in \mathbb{R}^{N' \times 64}$ with a dense spatial resolution of $L \times H \times W$. And we project the $F^0_v$ along the z-axis to generate the BEV feature $F_b \in \mathbb{R}^{64 \times H \times W}$. After that, the sparse 3D CNN, which consists of four encoder blocks used in GASN \cite{ye2022efficient} extracts the multi-scale 3D features $(F^1_v, F^2_v, F^3_v, F^4_v)$. While the 2D U-Net consisting of a stack of 2D convolutions encodes the BEV features $F'_b$ under different receptive fields. We further back-project the encoded 3D features, and BEV features to get the point-wise features $(f^0_p , f^1_p, f^2_p, f^3_p, f^4_p, f^b_p)$, which are concatenated along channel dimension and fed into MLPs for the fused features $f_p \in \mathbb{R}^{N \times 64}$. Finally, the semantic head and instance branch take the fused features $f_p$ as the input and output semantic confidences and instance proposals.

\textbf{Semantic head.} The MLPs are exploited to predict the semantic scores for each point. The cross-entropy and lovasz loss \cite{berman2018lovasz} are used to train the semantic head. Moreover, following \cite{ye2022efficient}, we use deep sparse supervision to accelerate the network's coverage.

\textbf{Instance branch.} As shown in Fig. \ref{fig:pipeline}(a), the instance branch consists of two modules, namely Sparse Instance Proposal (SIP) and Instance Aggregation (IA), which are further introduced in the following sections. We do not use the commonly used offset branch in \cite{hong2021lidar,zhou2021panoptic, liu2022prototype, li2022PHNet,duerr2022rangebird} to predict the geometric center shifts, which are used for subsequent instance grouping with clustering algorithms. we utilize the SIP module to directly generate instance proposals from the raw points of things. Moreover, we use the IA module to merge proposals of the same instance ID by the instance affinities to boost the SIP module's performance on large objects.
\subsection{Sparse Instance Proposal} \label{Sec.SIP}
The sparse instance proposal module receives the semantic predictions and raw point cloud as the input and generates the instance proposals, as shown in Fig. \ref{fig:pipeline}. Since directly grouping raw points of things will bring a large computational burden and memory overhead, we first introduce a Balanced Point-Sampling (BPS) strategy to obtain sparseS seed points.

\textbf{Balanced point-sampling.} The widely employed farthest point sampling (FPS) is computationally expensive. For example, it takes over 200 seconds to sample 10\% of 1 million points. While random sampling usually suffers from long-tail distribution problems \cite{cheng2022pcb}, i.e., the closer the distance to the sensor, the denser the point cloud. Therefore, inspired by PCB-RandNet \cite{cheng2022pcb}, we devise a Balanced Point-Sampling (BPS) strategy for generating sparse seed points of things. As shown in Fig.\ref{fig:pipeline}(b), the input point cloud is first divided into voxel blocks, similar to the voxelization operation in Sec. \ref{Sec.BD}. Then, we scatter the points in the same voxel and calculate the average of these points as the seed point. It means that each non-empty voxel contains one seed point that dominates all points in the same voxel block. Notably, our BPS is affected by the instance occupancy and voxel resolution rather than the point density. The seed point distribution over the distance range can be uniform through our BPS method.
Moreover, different from FPS and random sampling, which need to exploit K Nearest Neighbor (K-NN) to further assign corresponding thing points to seed points, our BPS is convenient and efficient to use the voxel block to implement the point sampling and assignment simultaneously.

\textbf{Bubble shrinking.} After the BPS, we can obtain the sparse seed points $X \in \mathbb{R}^{M \times 3}$ of things. 
The semantics of a seed point is generated by major voting on all points corresponding to the seed point. Furthermore, the shift module, termed bubble shrinking, is exploited to shift the seed points to centers of instances they belong to. We first give the minimum radius $r_c$ of thing category $c$ empirically. Then we establish a graph with all the sparse seed points as vertices. Two vertices are connected if they belong to the same category $c$ and their distance is smaller than the minimum radius $r_c$. We assign each seed point a bubble which contains points that connect to the seed point. Then similar to other shift modules \cite{comaniciu2002mean, hong2021lidar}, the bubbles are iteratively shrunk $L=4$ times according to the points in the bubble as shown in Fig.\ref{fig:pipeline}(c). In this way, the seed points will be more clustered. The procedure of bubble shrinking is illustrated in Algorithm \ref{alg:BS}. Our bubble shrinking is simple and efficient. Notably, in our shift module, the connectivity of the graph, i.e., the adjacency matrix $K$ is determined initially. There is no need to rebuild the connected graph in each iteration, reducing the computation and memory overhead. Moreover, we experimentally find that keeping the same graph connectivity across all iterations is more stable.

\textbf{Point grouping.} For instance proposals, we perform point grouping upon the shifted sparse seed points $X'$. As shown in Fig. \ref{fig:pipeline}(d), we use the connected component labeling to group points to instances. Similar to bubble shrinking, the shifted points are viewed as vertices in a graph, and the semantic categories and distances determine the connectivity among vertices. The distance thresholds are empirically set to half of the minimum radius in bubble shrinking. Then we regard the connected component in the graph as an instance.
Note that the points dominated by seed points in the same connected component share an instance ID. We define the grouped instance proposals as $\{I_i\}_{i=1}^{O}$. $O$ is the number of proposals.

\begin{algorithm}[t]
    \caption{Bubble Shrinking} \label{alg:BS}
    \SetAlgoLined
    \KwIn{Sparse seed points $X \in \mathbb{R}^{M\times3}$, Minimum radius $R=\{r_c\}_{c=1}^{N_c}$}
    \KwOut{Shifted seed points $X' \in \mathbb{R}^{M\times3}$}
    Construct the connected graph $G = (V, E)$ with adjacency matrix $K$; $(u, v) \in E$ if nodes $u,v$ belong to the same category $c$ and $L2(u, v) < r_c$\\
    $D = diag(K \mathbf{1}), X = X^T$ \\
    \For{$i \gets 1\ \KwTo\ L$} {
        $X=XD^{-1}K$
    }
    $X' = X^T$ \\
    \Return $X'$
\end{algorithm}	

\subsection{Instance Aggregation}
\begin{figure}[t]
\centering
	\includegraphics[width=0.46\textwidth]{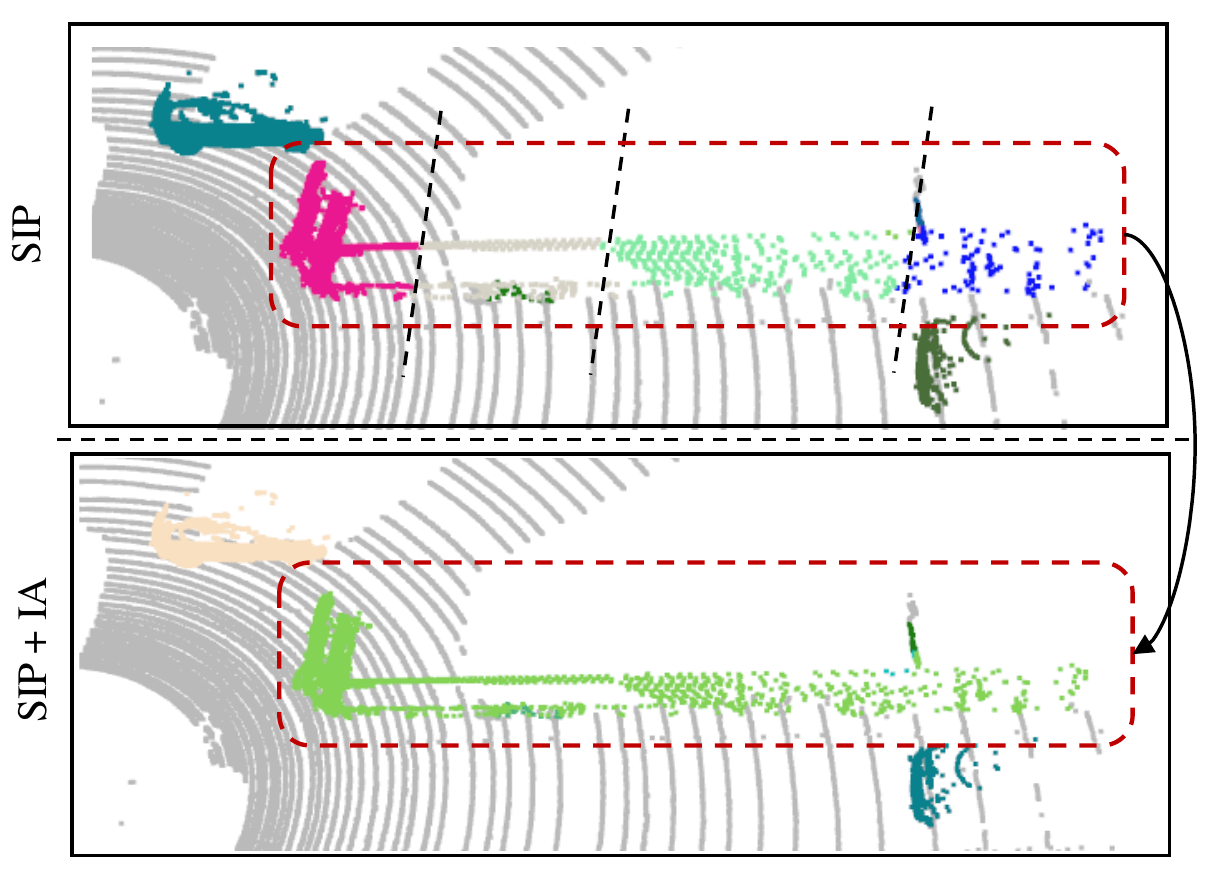}
	\caption{False cases where the bus is over-segmented by SIP module. And IA module can merge the fragmented instances effectively.}
	\label{fig:case}
\end{figure}

Due to the effective shifting and grouping, our SIP can reduce fragmented instances. However, we observed that it still suffers from the over-segment problem when grouping big objects such as buses and trucks due to the sparsity of LiDAR point cloud, as shown in Fig.\ref{fig:case}. To improve the performance on large objects, similar to GP-S3Net \cite{razani2021gp}, we propose to aggregate instances by the instance affinities. Specifically, given the points set $P_i$ and the point-wise features $F_i$ of $i$-th instance proposal, we utilize MLPs to enhance $F_i$ with the position $P_i$ and use the max pooling to aggregate the enhanced features into the global instance features $g_i$ of shape $[64]$. The procedure is formulated as follows:
\begin{equation}
    g_i = \mathrm{MaxPool}(\mathrm{MLP}([F_i, P_i]))
\end{equation} where $[\cdot]$ denotes concatenation.

We further apply the KNN-Transformer to implement the interactions among the global instance features $\{g_i\}_{i=1}^{O}$. Let $p_i = \rm{AvgPool}(P_i)$ is the center of $i$-th instance. The indices
of $K$ nearest neighbors for $p_i$ are calculated based on their spatial location. Then, we index the corresponding $K$ instance features $\{g_j\}_{j\in N(i)}$ according to the indices. Through linear transformations, $g_i$ is mapped to $q_i$ with shape $[64]$ and ($k_i, v_i$) with shape $[K, 64]$ is generated by $\{g_j\}_{j\in N(i)}$. Afterwards, the similarity weights between $q_i$ and $k_i$ are computed and multiplied with $v_i$ to obtain features $\hat{g}_i$: 
\begin{equation}
    \hat{g}_i = \mathrm{softmax}(\frac{q_i\circ k_i}{\sqrt{C}}) \circ v_i 
\end{equation} where $\circ$ denotes dot-product and $C=64$.

The enhanced global instance features $\{\hat{g}_i\}_{i=1}^{O}$ are exploited to calculate the instance affinities. The instance affinity $s_{i,j}$ of proposals $i$ and $j$ is computed according to their global features and the distance of instance centers. The procedure can be formulated as follows:
\begin{equation}
    s_{i,j} = \mathrm{sigmoid}[\mathrm{MLP}([\hat{g}_i, \hat{g}_j, |p_i - p_j|])]
\end{equation} which is supervised by binary cross-entropy loss:
\begin{equation}
    L_{aff} = \sum_{i,j}[y_{i,j}log(s_{i,j}) + (1-y_{i,j})log(1-s_{i,j})]
\end{equation} where $y_{i, j}$ is set 1 if proposals $i$ and $j$ share the same instance ID else 0. Note that the instance ID for each proposal is determined by major voting.
In the inference stage, we merge two instances if their affinity exceeds a certain threshold. Similar to the point grouping in Sec. \ref{Sec.SIP}, the merging procedure can also be implemented by the CCL algorithm efficiently. 

\section{Experiments}
We conduct extensive experiments on two large-scale datasets SemanticKITTI \cite{behley2019semantickitti} and nuScenes \cite{fong2022panoptic} to evaluate our PANet.
\subsection{Datasets and Metrics}
\textbf{SemanticKITTI.} SemanticKITTI \cite{behley2019semantickitti}, a large-scale dataset for autonomous driving, is derived from the KITTI odometry dataset \cite{geiger2012we}, which collects 22 data sequences (10 for training, 11 for testing, and 1 for validation) with a 64-beams LiDAR sensor. It is the first dataset that presents the challenge of LiDAR-based panoptic segmentation. SemanticKITTI provides annotated point-wise labels in 20 classes for segmentation tasks, among which 8 classes are defined as things classes, and the rest are stuff classes.

\textbf{NuScenes.} The other large-scale dataset nuScenes \cite{fong2022panoptic} also releases the challenge of panoptic segmentation. It contains a wide diversity of urban scenes, including 850 scenes for training and validation and 150 for testing.
nuScenes uses a 32-beam LiDAR sensor to create annotations in 2 Hz, which contains 6 stuff classes and 10 thing classes out of 16 semantic classes.

\textbf{Metrics.} As defined in \cite{behley2021benchmark}, we take Panoptic Quality (PQ), Segmentation Quality (SQ), and Recognition Quality (SQ) as the evaluation metrics. The metrics for things and stuff are calculated separately, i.e., $\mathrm{PQ^{Th}, SQ^{Th}, RQ^{Th}}$ and $\mathrm{PQ^{St}, SQ^{St}, RQ^{St}}$. In addition, we also report $\mathrm{PQ^\dag}$ defined by swapping the PQ of each stuff class to IoU and then averaging over all classes. Mean IoU (mIoU) is also adopted to evaluate the quality of semantic segmentation.

\subsection{Implementation Details}
The voxelization space for the backbone is limited in $[[\pm48], [\pm48], [-3, 1.8]]$ and the voxelization resolution is set to $[0.2, 0.2, 0.1]$ in meters. During training, similar to \cite{hong2021lidar}, we apply data augmentation such as global scaling and random flipping on the input points of both datasets. Our model is trained for 50 epochs following \cite{ye2022efficient} for semantic segmentation and another 10 epochs for the instance aggregation module with a total batch size of 8 on 2 NVIDIA RTX 3090 GPUs. The minimum radius in the bubble-shrinking module is empirically given according to the average size of things in each class. Moreover, we use per-category histogram thresholding to determine the merging threshold in the instance aggregation module. 

\subsection{Comparison with the State-of-the-art.}
\begin{table}[t]
    \caption{LiDAR panoptic segmentation results on the validation set of SemanticKITTI. SIP denotes Sparse Instance Proposal and Sem indicates semantic segmentation. All results in [\%].}
    \centering
    \setlength{\tabcolsep}{0.0195\linewidth}
    \begin{tabular}{l|cccc|c} \toprule
        Method & PQ & $\mathrm{PQ^\dag}$ & RQ & SQ & mIoU \\
        \midrule
        KPConv \cite{thomas2019kpconv} + PV-RCNN \cite{shi2020pv} & 51.7 & 57.4 & 63.1 & 78.9 & 63.1 \\
        PointGroup \cite{jiang2020pointgroup} & 46.1 & 54.0 & 56.6 & 74.6 & 55.7 \\
        LPASD \cite{milioto2020lidar} & 36.5 & 46.1 & - & - & 50.7 \\
        EfficientLPS \cite{sirohi2021efficientlps} & 59.2 & 65.1 & 69.8 & 75.0 & 64.9 \\
        Panoptic-PolarNet \cite{zhou2021panoptic} & 59.1 & 64.1 & 70.2 & 78.3 & 64.5 \\
        DSNet \cite{hong2021lidar} & 57.7 & 63.4 & 68.0 & 77.6 & 63.5 \\
        SCAN \cite{xu2022sparse} & 57.2 & - & - & - & \textbf{68.9} \\
        Panoptic-PHNet \cite{li2022PHNet} & \textbf{61.7} & - & - & - & 65.7 \\
        MaskPLS-M \cite{marcuzzi2023mask} & 59.8 & - & 69.0 & 76.3 & 61.9 \\
        \midrule
        Sem + Offset + MeanShift \cite{comaniciu2002mean} & 59.8 & 64.7 & 70.6 & 78.1 & 68.1 \\
        Sem + Offset + DBScan \cite{ester1996density} & 59.6 & 64.5 & 70.2 & 78.5 & 68.1 \\
        Sem + Offset + HDBScan \cite{campello2013density} & 60.0 & 64.9 & 70.4 & 78.7 & 68.1 \\
        Sem + Offset + DS \cite{hong2021lidar} & 60.1 & 65.0 & 70.7 & 78.5 & 68.1 \\
        Sem + MeanShift \cite{comaniciu2002mean} & 57.7 & 62.7 & 69.1 & 76.7 & 68.1 \\
        Sem + SIP (Ours) & 61.5 & 66.4 & 71.7 & 79.1 & 68.1 \\
        PANet (Ours) & \textbf{61.7} & \textbf{66.6} & \textbf{71.8} & \textbf{79.3} & 68.1 \\ \bottomrule
    \end{tabular}
    \label{tab:semkitti_val}
\end{table}

\begin{table*}[t]
    \vspace{-0.2cm}
    \caption{LiDAR panoptic segmentation results on the test set of SemanticKITTI. All results in [\%].}
    \centering
    \setlength{\tabcolsep}{0.0148\linewidth}
    \begin{tabular}{l|cccc|ccc|ccc|c} \toprule
        Method & PQ & $\mathrm{PQ^\dag}$ & RQ& SQ & $\mathrm{PQ^{Th}}$ &$\mathrm{RQ^{Th}}$ & $\mathrm{SQ^{Th}}$ & $\mathrm{PQ^{St}}$ &$\mathrm{RQ^{St}}$ & $\mathrm{SQ^{St}}$ & mIoU \\
        \midrule
        RangeNet++ \cite{milioto2019rangenet++} + PointPillars \cite{lang2019pointpillars} & 37.1 & 45.9 & 47.0 & 75.9 & 20.2 & 25.2 & 75.2 & 49.3 & 62.8 & 76.5 & 52.4 \\
        LPSAD \cite{milioto2020lidar} & 38.0 & 47.0 & 48.2 & 76.5 & 25.6 & 31.8 & 76.8 & 47.1 & 60.1 & 76.2 & 50.9 \\
        KPConv \cite{thomas2019kpconv} + PointPillars \cite{lang2019pointpillars} & 44.5 & 52.5 & 54.4 & 80.0 & 32.7 & 38.7 & 81.5 & 53.1 & 65.9 & 79.0 & 58.8 \\
        Panoster \cite{gasperini2021panoster} & 52.7 & 59.9 & 64.1 & 80.7 & 49.4 & 58.5 & 83.3 & 55.1 & 68.2 & 78.8 & 59.9 \\
        Panoptic-PolarNet \cite{zhou2021panoptic} & 54.1 & 60.7 & 65.0 & 81.4 & 53.3 & 60.6 & 87.2 & 54.8 & 68.1 & 77.2 & 59.5 \\
        CPSeg \cite{li2021cpseg} & 57.0 & 63.5 & 68.8 & 82.2 & 55.1 & 64.1 & 86.1 & 58.4 & 72.3 & 79.3 & 62.7 \\
        DS-Net \cite{hong2021lidar} & 55.9 & 62.5 & 66.7 & 82.3 & 55.1 & 62.8 & 87.2 & 56.5 & 69.5 & 78.7 & 61.6 \\
        EfficientLPS \cite{sirohi2021efficientlps} & 57.4 & 63.2 & 68.7 & 83.0 & 53.1 & 60.5 & 87.8 & 60.5 & 74.6 & 79.5 & 61.4 \\
        Panoptic-PHNet \cite{li2022PHNet} & \textbf{61.5} & 67.9 & 72.1 & 84.8 & 63.8 & 70.4 & 90.7 & 59.5 & 73.3 & 80.5 & \textbf{66.0} \\
        MaskPLS-M \cite{marcuzzi2023mask} & 58.2 & 69.3 & 68.6 & 83.9 & 55.7 & 61.7 & 89.2 & 60.0 & 73.7 & 80.0 & 62.5 \\
        \midrule
        PANet (Ours) & 58.5 & 65.2 & 69.8 & 83.0 & 59.7 & 68.1 & 87.2 & 57.6 & 71.0 & 79.9 & 64.6 \\ \bottomrule
    \end{tabular}
    \label{tab:semkitti_test}
\end{table*}

\begin{table*}[t]
    \vspace{-0.2cm}
    \caption{LiDAR panoptic segmentation results on the validation set of nuScenes. All results in [\%].}
    \centering
    \setlength{\tabcolsep}{0.018\linewidth}
    \begin{tabular}{l|cccc|ccc|ccc|c} \toprule
        Method & PQ & $\mathrm{PQ^\dag}$ & RQ& SQ & $\mathrm{PQ^{Th}}$ &$\mathrm{RQ^{Th}}$ & $\mathrm{SQ^{Th}}$ & $\mathrm{PQ^{St}}$ &$\mathrm{RQ^{St}}$ & $\mathrm{SQ^{St}}$ & mIoU \\
        \midrule
        PanopticTrackNet \cite{hurtado2020mopt} & 51.4 & 56.2 & 63.3 & 80.2 & 45.8 & 55.9 & 81.4 & 60.4 & 75.5 & 78.3 & 58.0 \\
        DS-Net \cite{hong2021lidar} & 42.5 & 51.0 & 50.3 & 83.6 & 32.5 & 38.3 & 83.1 & 59.2 & 70.3 & 84.4 & 70.7 \\
        EfficientLPS \cite{sirohi2021efficientlps} & 62.0 & 65.6 & 73.9 & 83.4 & 56.8 & 68.0 & 83.2 & 70.6 & 83.6 & 83.8 & 65.6 \\
        Panoptic-PolarNet \cite{zhou2021panoptic} & 63.4 & 67.2 & 75.3 & 83.9 & 59.2 & 70.3 & 84.1 & 70.4 & 83.5 & 83.6 & 66.9 \\
        GP-S3Net \cite{razani2021gp} & 61.0 & 67.5 & 72.0 & 84.1 & 56.0 & 65.2 & 85.3 & 66.0 & 78.7 & 82.9 & 75.8 \\
        PVCL \cite{liu2022prototype} & 64.9 & 67.8 & 77.9 & 81.6 & 59.2 & 72.5 & 79.7 & 67.6 & 79.1 & 77.3 & 73.9 \\
        SCAN \cite{xu2022sparse} & 65.1 & 68.9 & 75.3 & \textbf{85.7} & 60.6 & 70.2 & 85.7 & \textbf{72.5} & \textbf{83.8} & \textbf{85.7} & \textbf{77.4} \\
        MaskPLS-M \cite{marcuzzi2023mask} & 57.7 & 60.2 & 66.0 & 71.8 & 64.4 & 73.3 & 84.8 & 52.2 & 60.7 & 62.4 & 62.5 \\
        \midrule
        PANet (Ours) & \textbf{69.2} & \textbf{72.9} & \textbf{80.7} & 85.0 & \textbf{69.5} & \textbf{79.3} & \textbf{86.7} & 68.7 & 82.9 & 82.1 & 72.6 \\ \bottomrule
    \end{tabular}
    \label{tab:nuscenes_val}
\end{table*}

\textbf{Quantitative Results.} Table \ref{tab:semkitti_val} shows that PANet outperforms all baseline methods on SemanticKITTI validation by a large margin. The PANet performs slightly lower on mIOU than SCAN while surpassing SCAN by 4.5\% in terms of PQ. PANet also achieves competitive performance on the SemanticKITTI test set, as shown in Table \ref{tab:semkitti_test}. For example, PANet outperforms the clustering-based methods Panoptic-PolarNet, DS-Net, and EfficientLPS by 6.4\%, 4.6\% and 6.6\% in terms of $\mathrm{PQ^{Th}}$. Notably, the SIP module in our PANet requires no extra training. However, PANet still performs best on validation and outperforms most clustering methods on test split, demonstrating the effectiveness of our methods. We also evaluate PANet on nuScenes validation. The results are presented in Table \ref{tab:nuscenes_val}. Our approach achieves state-of-the-art performance on all reported metrics, which confirms the advantages of our PANet.

\begin{figure*}[t]
\centering
	\includegraphics[width=\textwidth]{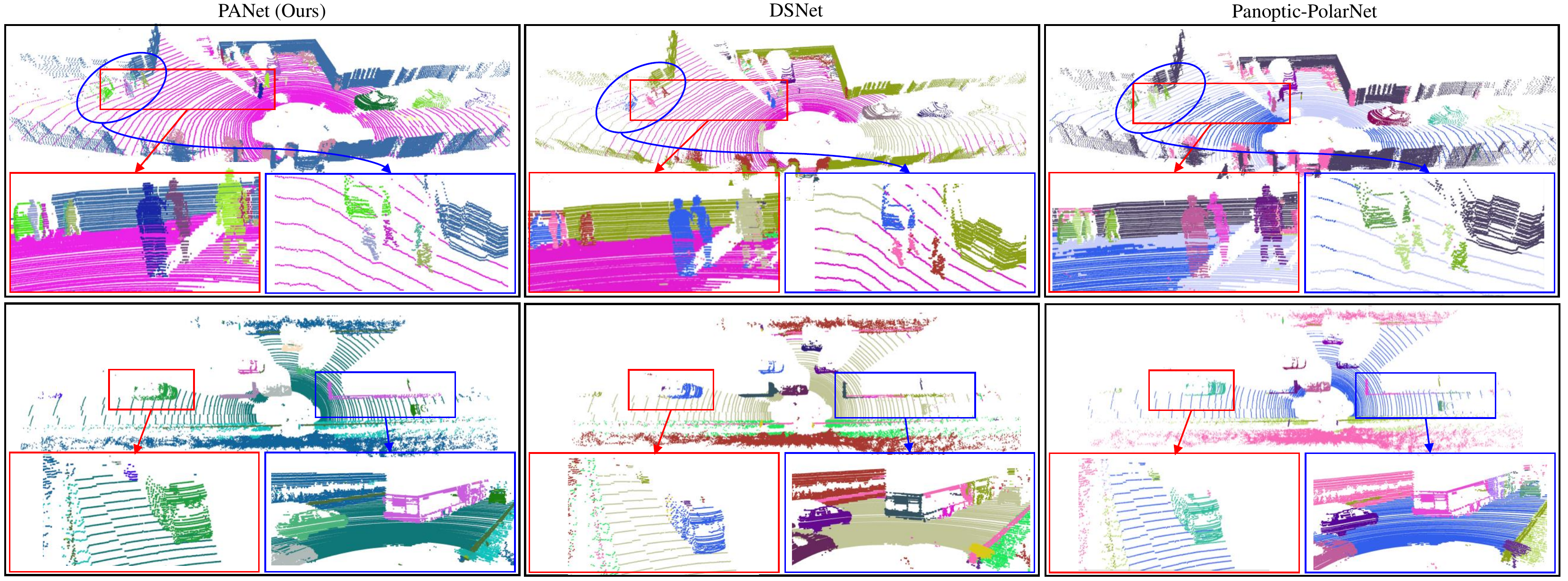}
	\caption{Visual comparison among DSNet, Panoptic-PolarNet, and PANet (Ours) on SemanticKITTI test split. Our PANet can handle crowded scenes and segment big objects correctly.}
	\label{fig:vis}
\end{figure*}

\textbf{Qualitative Results.} We visualize the results of our PANet and some LPS methods (DSNet \cite{hong2021lidar}, and Panoptic-PolarNet \cite{zhou2021panoptic}) with released codes on the SemanticKITTI test set to validate our PANet. As shown in Fig. \ref{fig:vis}, our PANet performs better than DSNet and Panoptic-PolarNet on both crowded scenes (upper part of the figure) and large object segmentation such as trucks and buses (bottom half).

\subsection{Ablation Study}
We conduct ablation studies on network components, clustering algorithms, and sampling algorithms on SemanticKITTI validation. The running time is tested on a single NVIDIA 1080 TI. 

\textbf{Ablation on network components.}
We analyze the effects of the 2D features, Sparse Instance Proposal (SIP) module, and Instance Aggregation (IA) module. Table \ref{ab:1} shows the detailed ablation studies. When taking the balanced point-sampling (BPS) and CCL-based point grouping (PG) for the instance proposal (line 1), our model has already achieved good performance, demonstrating the effectiveness of the combination of our BPS and CCL-based point grouping. Moreover, the bubble shifting (BS) can further boost the performance by 1.5\% on $\mathrm{PQ^{Th}}$ (line 1 vs. line 2).
Besides, the 2D features bring the gain by 0.6\% on PQ and 0.4\% on mIoU (line 2 vs. line 3). 
Our IA module also improves the performance by 0.5\% on $\mathrm{PQ^{Th}}$ (line 3 vs. line 4). 
As shown in Table \ref{ab:2}, IA presents a significant quality improvement on large objects such as Truck (+3.1\% on PQ), which shows its effectiveness in aggregating the over-segmented large objects.

We also provide another baseline to 
show the effectiveness of our non-learning SIP module. 
The baseline denoted as ``sem + MeanShift" comes from the combination of semantic predictions and 
MeanShift \cite{comaniciu2002mean} clustering on raw points of things, as shown in Table \ref{tab:semkitti_val}. 
Compared to the baseline, our SIP brings a significant improvement (+3.8\% on PQ). 

\newcommand{\tabincell}[2]{\begin{tabular}{@{}#1@{}}#2\end{tabular}} 
\begin{table}[t]
    \caption{Effect of the network components on the validation set of SemanticKITTI. BPS, BS, and PG denote Balanced point-sampling, Bubble shrinking, and Point grouping.}
    \centering
    \setlength{\tabcolsep}{0.0148\linewidth}
    \begin{tabular}{cccc|cc|cc|c} \toprule
        3D & 2D & \tabincell{c}{SIP \\ BPS / BS / PG} & IA & PQ & RQ & $\mathrm{PQ^{Th}}$ & $\mathrm{RQ^{Th}}$ & mIoU \\
        \midrule
        $\checkmark$ & $\times$ & $\checkmark$ / $\times$ / $\checkmark$ & $\times$ & 60.2 & 70.4 & 63.6 & 70.4 & 67.7 \\
        $\checkmark$ & $\times$ & $\checkmark$ / $\checkmark$ / $\checkmark$ & $\times$ & 60.9 & 71.0 & 65.1 & 71.9 & 67.7 \\
        $\checkmark$ & $\checkmark$ & $\checkmark$ / $\checkmark$ / $\checkmark$ & $\times$ & 61.5 & 71.7 & 67.0 & 74.4 & 68.1 \\
        $\checkmark$ & $\checkmark$ & $\checkmark$ / $\checkmark$ / $\checkmark$ & $\checkmark$ & \textbf{61.7} & \textbf{71.8} & \textbf{67.5} & \textbf{74.6} & 68.1 \\ \bottomrule
    \end{tabular}
    \label{ab:1}
\end{table}

\begin{table}[t]
    \caption{Per-class PQ results of IA module on the validation of SemanticKITTI. OV denotes Other-vehicle.}
    \centering
    \setlength{\tabcolsep}{0.045\linewidth}
    \begin{tabular}{c|c|cccc} \toprule
        Module & $\mathrm{PQ^{Th}}$ & Car & Truck & OV  \\
        \midrule
        w/o IA & 67.0 & 91.6 & 61.2 & 59.2 \\
         w/ IA & \textbf{67.5} & \textbf{92.1} & \textbf{64.3 (+3.1)} & \textbf{59.6} \\ \bottomrule
    \end{tabular}
    \label{ab:2}
\end{table}

\textbf{Ablation on clustering algorithms.}
We compare our PANet (the scheme of SIP attached with the IA) 
with three widely used clustering algorithms: MeanShift \cite{comaniciu2002mean}, DBScan \cite{ester1996density}, and HDBScan \cite{campello2013density}. We also give the results of the Dynamic Shift (DS) module \cite{hong2021lidar}.
Following common practices, we add an offset head composed of three linear layers to predict the point-wise offset vector to the instance center. 
And these clustering algorithms group instances upon the shifted points of things.
The results are presented in Table \ref{tab:semkitti_val} and show that our methods surpass all listed clustering methods in PQ accuracy. Significantly our PANet outperforms the DS module by 1.6\% in terms of PQ. Notably, the performance of MeanShift clustering without the offset branch decreases by 2.1\% in terms of PQ, which further clarifies our claim in Sec. \ref{Sec.intro}.
Furthermore, the model equipped with only our SIP achieves 61.5\% in terms of PQ and performs better than those baselines which exploit offset head for the geometric shift. We also provide a detailed comparison between our SIP and DS modules in Table \ref{ab:4}, showing that our model performs better in both accuracy and speed ($\sim13\times$ faster).
Notably, our SIP is non-learning and can be extended to 
other backbones and datasets in a plug-and-play manner.  

\begin{table}[t]
    \caption{Comparison between SIP and DS modules.}
    \centering
    \setlength{\tabcolsep}{0.035\linewidth}
    \begin{tabular}{c|ccc|c} \toprule
        Module & $\mathrm{PQ^{Th}}$ & $\mathrm{RQ^{Th}}$ & $\mathrm{SQ^{Th}}$ & Time (ms) \\
        \midrule
        DS \cite{hong2021lidar} & 63.8 & 72.1 &  85.9 & 259.1  \\
        SIP (Ours) & \textbf{67.0} & \textbf{74.6} & \textbf{87.2} & \textbf{19.5} \\
        \bottomrule
    \end{tabular}
    \label{ab:4}
\end{table}

\begin{table}[t]
    \caption{Ablation on point-sampling algorithms.}
    \centering
    \setlength{\tabcolsep}{0.035\linewidth}
    \begin{tabular}{c|ccc|c} \toprule
        Module & $\mathrm{PQ^{Th}}$ & $\mathrm{RQ^{Th}}$ & $\mathrm{SQ^{Th}}$ & Time (ms) \\
        \midrule
        FPS & 67.5 & 74.3 & 87.6 & 58.8   \\
        Random & 66.6 & 74.2 & 87.1 & 36.3  \\
        BPS (Ours) & 67.5 & 74.6 & 87.6 & \textbf{19.5} \\
        \bottomrule
    \end{tabular}
    \label{ab:3}
\end{table}

\textbf{Ablation on point-sampling algorithms.}
We replace the Balanced point-sampling (BPS) module with the farthest 
point-sampling (FPS) and random sampling algorithms to further 
demonstrate the effectiveness of our BPS.
For a fair comparison, the number of seed points is 
consistent for the three methods. And for the FPS and random sampling algorithms, the K Nearest Neighbor (K-NN) algorithm is applied to assign the corresponding points to the seed points. For convenience,  we test the running time of SIP modules equipped with different sampling algorithms to provide an efficiency comparison. The results are listed in Table \ref{ab:3} and show that our BPS achieves comparable performance with FPS in accuracy and has the lowest latency. For example, BPS outperforms random sampling by 0.9\% in terms of $\mathrm{PQ^{Th}}$ and is 3 times faster than FPS. 
It is noteworthy that random sampling is slower than our BPS. 
The reason lies in the K-NN assignment, which brings the additional computation overhead.



\section{Conclusion}
In this paper, we propose a LiDAR panoptic segmentation framework named PANet to avoid the dependency on the offset branch and the over-segmented problem on large objects. PANet devises a non-learning Sparse Instance Proposal (SIP) module to directly group thing points into instances from the raw point cloud efficiently. SIP does not require extra training and can be easily extended to other backbones and datasets in a plug-and-play manner. Moreover, the Instance Aggregation (IA) module is proposed to integrate the fragmented instances, which can complement the SIP module and improve the completeness of segmentation on large objects. Our method achieves the state-of-the-art among published works on both SemanticKITTI and nuScenes benchmarks.

\section*{ACKNOWLEDGMENT}
This work was supported by NSFC 62088101 Autonomous Intelligent Unmanned Systems.

\bibliographystyle{IEEEtran}
\bibliography{IEEEabrv,ref}


\end{document}